\documentclass[letterpaper]{article} % DO NOT CHANGE THIS
\usepackage{aaai24}  % DO NOT CHANGE THIS
\usepackage{times}  % DO NOT CHANGE THIS
\usepackage{helvet}  % DO NOT CHANGE THIS
\usepackage{courier}  % DO NOT CHANGE THIS
\usepackage[hyphens]{url}  % DO NOT CHANGE THIS
\usepackage{graphicx} % DO NOT CHANGE THIS
\usepackage{natbib}  % DO NOT CHANGE THIS AND DO NOT ADD ANY OPTIONS TO IT
\usepackage{caption} % DO NOT CHANGE THIS AND DO NOT ADD ANY OPTIONS TO IT
\usepackage{booktabs}       % professional-quality tables
\usepackage{amsfonts}       % blackboard math symbols
\usepackage{xcolor}         % colors
\usepackage{amsmath}
\usepackage{mathrsfs}
\usepackage{bm}
\usepackage{algorithm}
\usepackage{algorithmic}

\usepackage{multirow}

\usepackage{subfig}

\usepackage{array}
\usepackage{url}
\usepackage{dsfont}
\usepackage{makecell}
\usepackage{enumitem}
\usepackage{svg}

\usepackage{newfloat}
\usepackage{listings}
\DeclareCaptionStyle{ruled}{labelfont=normalfont,labelsep=colon,strut=off} % DO NOT CHANGE THIS
\floatstyle{ruled}
\newfloat{listing}{tb}{lst}{}
\floatname{listing}{Listing}
\title{Full Bayesian Significance Testing for Neural Networks}
\author {
    Zehua Liu\textsuperscript{\rm 1},
    Zimeng Li\textsuperscript{\rm 1},
    Jingyuan Wang\textsuperscript{\rm 1,\rm 2,\rm 3}\thanks{Corresponding author (jywang@buaa.edu.cn)},
    Yue He\textsuperscript{\rm 4}
}
\affiliations {
    \textsuperscript{\rm 1}School of Computer Science and Engineering, Beihang University, Beijing, China\\
    \textsuperscript{\rm 2}School of Economics and Management, Beihang University, Beijing, China\\
    \textsuperscript{\rm 3}Key Laboratory of Data Intelligence and Management (Beihang University), \\Ministry of Industry and Information Technology, Beijing, China\\
    \textsuperscript{\rm 4}Department of Computer Science and Technology, Tsinghua University, Beijing, China
}
\usepackage{bibentry}
\begin{document}

\maketitle

\begin{abstract}
Significance testing aims to determine whether a proposition about the population distribution is the truth or not given observations.
However, traditional significance testing often needs to derive the distribution of the testing statistic, failing to deal with complex nonlinear relationships.
In this paper, we propose to conduct Full Bayesian Significance Testing for neural networks, called \textit{n}FBST, to overcome the limitation in relationship characterization of traditional approaches.
A Bayesian neural network is utilized to fit the nonlinear and multi-dimensional relationships with small errors and avoid hard theoretical derivation by computing the evidence value.
% , a function of the posterior distributions of the null hypothesis and the alternative.
%, with a theoretical convergence.
%\textbf{We also theoretically prove that the evidence value of a sharp null hypothesis goes to one as the sample size goes to infinite.}
Besides, \textit{n}FBST can test not only global significance but also local and instance-wise significance, which previous testing methods don't focus on.
Moreover, \textit{n}FBST is a general framework that can be extended based on the measures selected, such as Grad-\textit{n}FBST, LRP-\textit{n}FBST, DeepLIFT-\textit{n}FBST, LIME-\textit{n}FBST. 
A range of experiments on both simulated and real data are conducted to show the advantages of our method. 
\end{abstract}

\section{Introduction}

% 引入显著性检验，但标准的检验假设很强，绕不开复杂的理论推导，因此采用FBST
Significance testing aims to determine whether a proposition about the population distribution\footnotemark[1] is true or false given observations, which is widely used in many scientific fields, such as social sciences\cite{orlitzky2012can, ortega2017bayesian} and medical research\cite{matthews1990analysis, rutledge2004effect}.
For example, it is often used to evaluate the efficacy of new treatments or drugs.
First, clinical trials are performed to compare the response of patients treated with a new therapy against a control group.
Then, significance testing is used as an analytical tool to determine whether the observed improvement in the treatment group is significant, which provides evidence that the new therapy is effective.

\footnotetext[1]{https://online.stat.psu.edu/stat462/node/249/}

To attain a proper testing result, a golden standard is to recover the true data generation model $f_0$ behind the population distribution, then justify the proposition according to $f_0$ directly.
% For this purpose, a number of significance testing approaches are proposed under the assumption that the correlations between the variables are linear.  
% However, the linear assumption is ideal beyond the practice. 
% The complicated nonlinear correlations abound in real applications. 
For this purpose, a number of significance testing approaches are proposed under different assumptions about $f_0$ \cite{gozalo1993consistent, lavergne1996nonparametric, racine1997consistent}.
However, simple assumptions can hardly fit the real situation, while complex assumptions make it hard to derive the theoretical distribution of the testing statistic.
Recently, \cite{horel2020significance} provides a provable solution of significance testing in nonlinear cases, but it suffers from the computational difficulty in statistics, only addressing a limited function space. 
There is still a gap to be solved for significance testing in the general correlations. 

% From another perspective, 
% traditional significance test only considers the global proposition under the linear assumption.
Existing significance testing methods only focus on global propositions.
However, some propositions that are invalid globally can still hold on and contribute to a certain sub-population.
For example, clinical trials have shown that a drug is effective in treating cancer, but not in some individuals.
For a better justification in nonlinear cases, a significance testing approach should verify the correctness of a proposition in population distribution and sub-population distribution respectively.

To deal with the complicated real data in wide applications\cite{he2020learning, wang2018multilevel, wang2020deep, wang2023wavelet}, we introduce deep neural networks into the significance testing to capture the nonlinear correlations.
To overcome the barrier of computing statistics under the complex fitting functions, we solve the significance testing problem from the Bayesian perspective \cite{kass1995bayes}, and propose a novel approach that conducts the Full Bayesian Significance Testing for neural networks, abbreviated as \textit{n}FBST (neural FBST).
Given the testing statistics, \textit{n}FBST can test the correctness of a proposition for both population-level or sub-population-level problems, by comparing the posterior probabilities of it and its opposite. 
In addition, \textit{n}FBST is a general framework that can be extended based on different testing statistics, such as Grad-\textit{n}FBST, LRP-\textit{n}FBST, DeepLIFT-\textit{n}FBST, LIME-\textit{n}FBST, and so on. A range of experiments on both simulated and real data are conducted to show the advantages of our method. 
The main contributions can be summarized as follows:
\begin{itemize}[leftmargin=15pt]
    \item We are the first to introduce deep neural networks into significance testing.
    % and solve it from the Bayesian perspective.
    % The classical testing methods rely on the specific assumptions of $f_0$, which limits their applicability.
    Our approach replaces complicated theoretical derivation by fitting distributions in a Bayesian way, and the neural network serves as a good estimator of $f_0$ without assuming specific forms.
  % \item We aim to explore knowledge hidden behind the underlying relationships between features and targets in a rigorous way.
  % It offers a new perspective distinct from most previous research that explains the model trained on the data.
  % To the best of our knowledge, we are the first to introduce significance testing into deep neural networks.
    \item We design a complete procedure using Full Bayesian Significance Testing for Neural Networks, namely \textit{n}FBST.
    It is a general framework that can be extended based on different implementations and different testing statistics, such as Grad-\textit{n}FBST, LRP-\textit{n}FBST, DeepLIFT-\textit{n}FBST, LIME-\textit{n}FBST, and so on.
    \item Our proposed \textit{n}FBST can solve both local and global significance testing problems while previous methods only focus on the latter. 
    % the testing problems in different dimensions.
    Under non-linear assumptions, global significance may be inconsistent with local or instance-wise significance.
    % However, existing testing methods only focus on the former and ignore the research on the latter.
    % Our approach can solve the global significance, as well as the local or instance-wise significance.
    \item We conduct extensive experiments to verify the advantage of our method on better testing results. 
    % The testing result obtained by \textit{n}FBST is not only statistically meaningful but also effectively distinguishes whether a feature is insignificant or not.
    % A range of experiments on both simulated and real data are conducted to show the advantages of our method. 
  % Further, we propose Q-GS, a Global Interpretability definition based on Quantile, which can extend the local interpretability method to the global judgment of the feature, and experiments show its validity.
  % \item We perform extensive experiments on both simulated and real data, including an empirical application to residential energy efficiency evaluation as well as MNIST.
  % We demonstrate that \textit{n}FBST performs better than previous methods and our method could discover accurate domain knowledge and provide insightful directions for future researches.
  % We analyze specifically why we need \textit{n}FBST and show the improvement before and after applying it.
  % Moreover, in some cases, existing methods could not correctly distinguish between significant and insignificant features.
  % Hence, our method provides a better understanding of these intriguing phenomena.
\end{itemize}

\section{Theoretical Method}

\subsection{Classical Frequentist Significance Test}\label{section classical significance test}

We denote $f_0: \mathcal{X} \subset \mathbb{R}^d \rightarrow \mathbb{R}$ as the underlying and unknown conditional mean function, namely $\rm E[Y\vert X=x]$, of the population $(X, Y)$.
% reflecting the data-generating process, 
% and $\theta$ be the parameters of $f_0$.
Then, we consider the data generation process of $(X, Y)$ as follows:
%\begin{small}
\begin{equation}\small
    y = f_0(X) + \epsilon,
\end{equation}
%\end{small}
where $\epsilon$ is a random error such that $\rm E[\epsilon|X] = \rm E[\epsilon] = 0$.
Significance testing first defines a testing statistic $\eta$, then proposes two contradictory propositions (or hypotheses) $H_0$ and $H_1$, which represent the null hypothesis and the alternative hypothesis respectively.
Classical significance testing is regarded as a procedure for measuring the consistency of data with the null hypothesis by the calculation of a \textit{p-value} (tail area under the null hypothesis) \cite{DeBragancaPereira1999}.
The process is as follows:
\begin{itemize}[leftmargin=15pt]
    \item First, we make assumptions about the population distribution of $f_0$ and denote it as $f_0(\beta)$, whose parameters are $\beta$.
    Then we derive the theoretical distribution of $\eta(\beta)$ under the assumptions.
    % \item Second, define a testing statistic $\eta(\theta)$, which is a function with respect to $\theta$. Then, 
    \item Second, based on the observed data $\mathcal{D}$, we fit an optimal estimator $\hat{f}(\hat{\beta})$ as an approximation function of $f_0$, whose parameters are $\hat{\beta}$.
    \item Third, we calculate $\eta(\hat{\beta})$ and \textit{p-value} further to determine whether the distribution of $\eta(\beta)$ is reasonable under the null hypothesis using sample information $\eta(\hat{\beta})$.
\end{itemize}
To test the significance of a feature, the problem is formulated as:
\begin{equation}\label{equation classical testing}\small
  H_0 : \eta(\beta) = 0    \quad
  H_1 : \eta(\beta) \neq 0,
\end{equation}
where $\eta(\beta)$ is a measure of feature importance. For example, if we assume $f_0$ satisfies linear relationships as follows:
\begin{equation}\small
    y = \beta_0 + \beta_1 x_1 + \dots + \beta_d x_d + \epsilon.    
\end{equation}
Whether the coefficient of a feature $x_j$ is equal to zero determines its significance, that is $\eta(\beta)=\beta_j$. 

However, there are two main defects in classical significance testing.
\begin{itemize}[leftmargin=15pt]
    \item First, the effectiveness of classical significance testing is based on reasonable assumptions about $f_0$, such as linear regression or kernel regression.
    However, it is difficult to find such precise assumptions when the data distribution is actually complicated.
    \item Second, some models, such as deep learning, excel in accurately fitting complex data distributions.
    However, the more complex assumption of $f_0$, the more computational theoretical distribution of $\eta(\beta)$, even intractable.
    % However, as the assumption of $f_0$ becomes more complex, the theoretical distribution of $\eta(\beta)$ is more computational, even intractable.
\end{itemize}

\subsection{Full Bayesian Significance Test}

In order to solve the problems, we adopt the Full Bayesian significance Testing (FBST) \cite{DeBragancaPereira1999, pereira2008can}.
FBST is a statistical methodology that allows for the testing of precise hypotheses in a Bayesian framework.
Here, ``full'' means that one only needs to use the posterior distribution to test without the specific assumptions for $f_0$.
In contrast to classical significance testing which uses \textit{p-value} to reject or fail to reject the null hypothesis, FBST provides a measure of evidence in favor of or against the null hypothesis, taking into account prior information and the strength of observations.

Let $P(H)$ be the prior probability of the hypothesis $H$ and $P(\mathcal{D} | H)$ be the likelihood function of $H$ given observations $\mathcal{D}$.
The posterior probability distributions for the null and alternative hypotheses are then calculated using Bayes' theorem as
\begin{equation}\small
\begin{aligned}
    % P(H_0 | \mathcal{D}) = \frac{P(\mathcal{D} | H_0) \times P(H_0)}{P(\mathcal{D})},
    % P(H_1 | \mathcal{D}) = \frac{P(\mathcal{D} | H_1) \times P(H_1)}{P(\mathcal{D})}.
    P(H | \mathcal{D}) = \frac{P(\mathcal{D} | H) \times P(H)}{P(\mathcal{D})} \propto {P(\mathcal{D} | H) \times P(H)}.
\end{aligned}
\end{equation}
% It states that $P(H | \mathcal{D})$ is proportional to the product of $P(H)$ and $P(\mathcal{D} | H)$.
It is consistent with the process by which people adjust their assessments in response to observed data.
The evidence in favor of the null hypothesis is quantified by the Bayes Factor 
\begin{equation}\small
    BF = \frac{P(\mathcal{D} | H_0) / P(\mathcal{D} | H_1)}{P(H_0) / P(H_1)}.
\end{equation}
% \begin{equation}\small
%     BF = P(\mathcal{D} | H_0) / P(\mathcal{D} | H_1).
% \end{equation}
Its value reflects which proposition is more likely under the observed data.
If it is greater than 1, we believe it provides evidence in favor of $H_0$.
The evidence is moderate if it is greater than $3$ and strong if it is greater than $10$ \cite{jeffreys1998theory}.
On the contrary, it provides evidence against $H_0$ if it is smaller than 1, moderate evidence for less than $\frac{1}{3}$, and strong evidence for less than $0.1$.

From the above analysis, we can conclude that FBST doesn't need to assume a specific distribution form of $f_0$, but calculate $P(\mathcal{D} | H_0)$ and $P(\mathcal{D} | H_1)$ instead.
In other words, the current goal is to obtain a good estimator to fit $P(\mathcal{D} | H)$.

\subsection{Approximate the Distribution of Testing Statistics}\label{section approximation the distribution of testing statistics}

According to the universal approximation theorem, neural networks with appropriate size can approximate an extensive class of functions to a desired degree of accuracy \cite{hornik1989multilayer}.
In this paper, we propose to use Bayesian neural networks to fit the likelihood $P(\mathcal{D} | H)$.
As a technique that combines Bayesian Theory and neural networks, Bayesian neural networks can fit complex relationships and produce a probability distribution over model parameters $\theta$ that expresses our beliefs regarding how likely the different parameter values are.

Given a dataset $\mathcal{D} = \{ (X^{(1)}, y^{(1}), \dots, (X^{(n)}, y^{(n)}) \}$, \textit{n}FBST first uses Bayesian neural networks, whose parameters are $\theta$, to fit $\mathcal{D}$.
Before training, a prior distribution is assigned to model parameters $\theta$ as an initial belief $\pi(\theta)$ according to experience.
This belief is gradually adjusted to fit data $\mathcal{D}$ by using the Bayesian rule.
The final belief is presented as the posterior distribution
\begin{equation}\label{the posterior of model parameters}\small
  P(\theta | \mathcal{D}) = \frac{P(\mathcal{D}|\theta)\pi(\theta)}{P(\mathcal{D})} 
  = \frac{\pi(\theta) \prod_{i=1}^n P(y^{(i)} | X^{(i)}, \theta)}{\int_\Theta \prod_{i=1}^n P(y^{(i)} | X^{(i)}, \theta) \mathrm{d}\theta} ,
\end{equation}
where $\Theta$ is the parameter space. Given a new case $X$, the prediction made by the Bayesian neural network is the weighted average of an ensemble
\begin{equation}\label{the posterior of y}\small
  P(y | X, \mathcal{D}) = \int_\Theta P(y | X, \theta) P(\theta | \mathcal{D}) \mathrm{d} \theta.
\end{equation}
Then, based on the posterior distribution of $\theta$, we obtain the posterior distribution of the testing statistic $\eta(\theta)$ and denote $p(\eta(\theta) | \mathcal{D})$ as its probability density.
The testing problem is formulated as:
\begin{equation}\label{equation fbst testing}
\small
    H_0 : \eta(\theta) = 0    \quad
    H_1 : \eta(\theta) \neq 0
\end{equation}
% \begin{figure}[t]
%     \centering
%     \includegraphics[width=0.5\columnwidth]{p_s_test.pdf}
%     \caption{Bayesian evidence calculated based on the distribution of $\eta(\theta)$.}
%     \label{fig_p_s_test}
% \end{figure}
\begin{figure}[t]
  \centering
  \includegraphics[width=0.6\columnwidth]{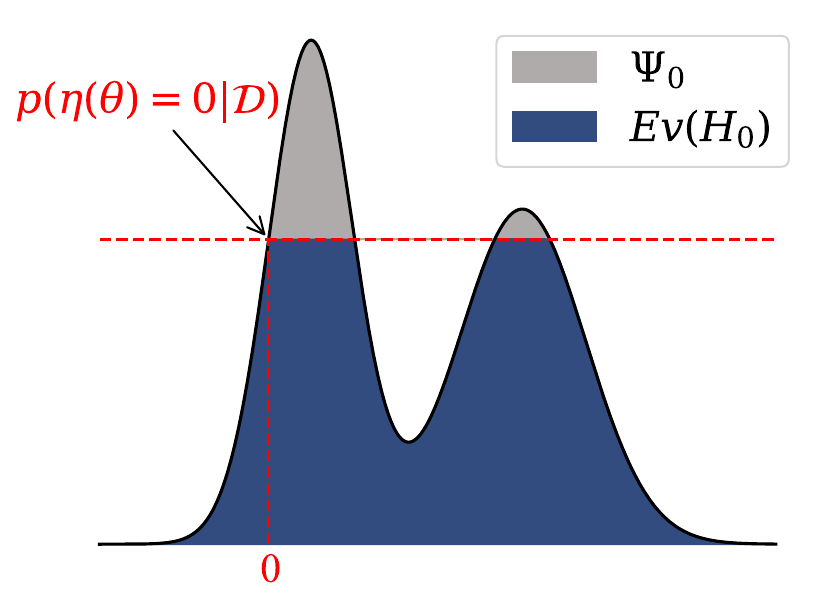}
  \caption{Bayesian evidence calculated based on the distribution of $\eta(\theta)$.}
  \label{fig_p_s_test}
\end{figure}
We denote the whole space of $\eta(\theta)$ as $\Psi$ such that $\eta(\theta) \in \Psi$.
Then, we define the region whose probability greater than $p(\eta(\theta)=0 | \mathcal{D})$ according to the following formula:
\begin{equation}\small
    \Psi_0=\{ \eta(\theta) : p(\eta(\theta) | \mathcal{D}) > p(\eta(\theta)=0 | \mathcal{D}) \},
\end{equation}
where $p(\eta(\theta)=0 | \mathcal{D})$ is the maximum of the posterior density under the null hypothesis $H_0$ (Figure \ref{fig_p_s_test}).
The Bayes Factor is not the only way to calculate evidence and its result is influenced by prior distribution.
In our method, we adopt a more flexible valid Bayesian evidence for the null hypothesis provided by \cite{DeBragancaPereira1999}:
\begin{equation}\label{equation_p_T0}\small
\begin{aligned}
    Ev(H_0)
    &= 1 - \int_{\Psi_0}{p \biggl( \eta(\theta) | \mathcal{D} \biggr)} \mathrm{d}\eta(\theta) \\
    % = P \biggl( \eta(\theta) \in T_0 | \mathcal{D} \biggr) 
    % = \int_{\Theta}{ \mathds{1} \biggl( \eta(\theta) \in T_{\eta^*(\theta)} \biggr) p \biggl( \eta(\theta) | \mathcal{D} \biggr)} \mathrm{d}\theta,
    % = 1 - \int_{\Theta}{ \mathds{1} \biggl( \eta(\theta) \in T_{0} \biggr) p \biggl( \eta(\theta) | \mathcal{D} \biggr)} \mathrm{d}\theta.
    &= 1 - \int_{\Psi}{ \mathds{1} \biggl( \eta(\theta) \in \Psi_{0} \biggr) p \biggl( \eta(\theta) | \mathcal{D} \biggr)} \mathrm{d}\eta(\theta).
\end{aligned}
\end{equation}
Using the Monte Carlo method, the above formula can be further simplified to
\begin{equation}\label{equation_p_T0_monte_carlo}\small
\begin{aligned}
    Ev(H_0)
    % = P \biggl( \eta(\theta) \in T_0 | \mathcal{D} \biggr) 
    &\approx 1 - \frac{1}{m} \sum_{i=1}^{m}{\mathds{1} \biggl( \eta_i(\theta) \in \Psi_{0} \biggr) } \\
    &= 1 - \frac{1}{m}\sum_{i=1}^{m}{\mathds{1}\biggl( p(\eta_i(\theta) | \mathcal{D}) > p(0 | \mathcal{D}) \biggr)},
\end{aligned}
\end{equation}
where $\eta_i(\theta)$ is sampled $m$ times based on the posterior probability density of $\eta(\theta)$.
The result of Eq (\ref{equation_p_T0_monte_carlo}) is called Bayesian evidence, whose value is between $0$ and $1$.
The closer the Bayesian evidence to $1$, the more likely to accept $H_0$.
The closer the Bayesian evidence to $0$, the more likely to reject $H_0$.
Moreover, we have mathematically proven that under certain constraints, as the sample size approaches infinity, $Ev(H_0)$ for insignificant features converges to $1$. 
The detailed process of proof is provided in the Appendix.

\section{Implementation Approach}

\subsection{Calculate the Distribution of Testing Statistics}

So far, we have clarified the entire process of FBST, but there still remain two implementation details that need to be elaborated on.
First, to perform \textit{n}FBST to deal with the testing problem Eq (\ref{equation fbst testing}), we need to calculate the posterior distribution of $\theta$.
Second, after obtaining the distribution of $\theta$, we need to calculate the distribution of the testing statistic $\eta(\theta)$.

% In other words, $P(\theta | \mathcal{D})$ is the foundation of the subsequent testing process.

In practice, it is intractable to solve the integral in Eq (\ref{the posterior of model parameters}).
A popular way, known as Variational Inference (VI), entails approximating the real but intractable posterior distribution with a tractable distribution called variational distribution \cite{blei2017variational,jaakkola2000bayesian}.
% A popular way, called variational inference , is to use a tractable distribution to approximate the real but intractable posterior distribution.
Therefore, Eq (\ref{the posterior of model parameters}) could be efficiently approximated.
Formally, variational family $Q = \{q_\vartheta: \vartheta \in \Gamma \}$ is a predefined family of tractable distributions on model parameter space $\Theta$, where $\vartheta$ is the parameter of variational distribution and $\Gamma$ is the range of $\vartheta$.
The optimal variational distribution $q_{\vartheta^*}$ is chosen from $Q$ such that
\begin{equation}\label{variational distribution}\small
\vartheta^{*} = \arg \min_{\vartheta \in \Gamma} \operatorname{KL}( q_\vartheta(\theta) \Vert P(\theta | \mathcal{D})) .
\end{equation}
KL divergence describes the ``distance'' between two distributions.
% Besides KL divergence \cite{graves2011practical,blundell2015weight}, $\alpha$ divergence \cite{li2017dropout} and Wasserstein distance \cite{ambrogioni2018wasserstein} are also utilized to measure the difference of distributions and learn the optimal approximation distribution to the real posterior.
We set diagonal Gaussian distributions as the prior and variational families of parameter $\theta$. 
This assumption is common in many works \cite{blundell2015weight, kendall2017uncertainties}.
Under this assumption, applying Bayesian rule Eq (\ref{the posterior of model parameters}), Eq (\ref{variational distribution}) can be further simplified as
\begin{equation}\label{simplified variational distribution}\small
\begin{aligned}
    \vartheta^* = \arg \min_{\vartheta \in \Gamma}-\mathbb{E}[\log{P(\mathcal{D} | \theta)}]+ \operatorname{KL}( q_\vartheta(\theta) \Vert \pi(\theta)) 
    +\log{P(\mathcal{D})} .
\end{aligned}
\end{equation}
The derivation is shown in Appendix.
The first term is related to data (such as MSE for regression task); The second term is only related to parameters $\theta$ like regularization term, and the third term is a constant.
% In the process of training Bayesian neural networks, we apply the reparameterization technique \cite{blundell2015weight}.
In the end, we finish approximating the posterior distribution of parameters $P(\theta | \mathcal{D})$ with variational distribution $q_{\vartheta^*}(\theta)$.

% The advantage of FBST is that we fit the distribution $P(\theta | \mathcal{D})$ without knowing its specific form.
% Therefore, we sample from $q_{\vartheta^*}(\theta)$ randomly and calculate corresponding values of the test statistic values to simulate its distribution.
We adopt Kernel Density Estimation (KDE) \cite{scott1979optimal, parzen1962estimation} to estimate the posterior probability density of $\eta(\theta)$, which is a common non-parametric method.
The process is as follows:
\begin{itemize}[leftmargin=15pt]
    \item First, draw samples of parameters $\theta$ with size $m$ from the approximate posterior distribution $q_{\vartheta^*}(\theta)$ randomly, that is $\{\theta_1 , \dots , \theta_m \}$.
    \item Second, calculate $\{ \eta(\theta_1), \dots, \eta(\theta_m) \}$ to obtain sample from the posterior of $\eta(\theta)$.
    \item Third, estimate $p(\eta(\theta) | \mathcal{D})$ using KDE 
    \begin{equation}\small
        p\biggl(\eta(\theta) | \mathcal{D}\biggr) = \frac{1}{mh}\sum_{i=1}^{m}{K\biggl(\frac{\eta(\theta) - \eta(\theta_i)}{h}\biggr)},
    \end{equation}
    where $K$ is the kernel function, and $h$ is the window width (also known as bandwidth).
    The commonly used kernel function is the Gaussian kernel function.
    % , $K(x) = \frac{1}{\sqrt{2\pi}} e^{-\frac{x^2}{2}}$.
\end{itemize}

Finally, by calculating Bayesian evidence in Eq (\ref{equation_p_T0_monte_carlo}), we finish the entire process of \textit{n}FBST.
\textit{n}FBST is a general and flexible framework that can be easily extended based on different implementations, including VI and KDE.
In the case of VI, we have derived the detailed training procedure in the Appendix.
The approximation between variational and posterior distributions can be gauged through the prediction error.
% We have derived the complete training procedure for variational inference in the Appendix.
% The degree of approximation between the variational and posterior distributions can be measured by the prediction error.
% From Table 2 in Appendix C, we can observe that the error is very close to the variance of generating data. Therefore, we can conclude that variational inference has approximated well.
As for KDE, its robust theoretical underpinning guarantees convergence and consistency, as evidenced by the work of \cite{parzen1962estimation}.
% Besides, there is a solid theoretical proof for the convergence and consistency \cite{parzen1962estimation} for KDE.
% Therefore, the error of our approach is within a reasonable range.
 Consequently, the margin of error in our approach remains within a reasonable range.

\subsection{Design of Testing Statistics}\label{section design of testing statistics}

% For \textit{n}FBST, the testing statistic $\eta(\theta)$ is defined as a function of parameters $\theta$.
% Under linear assumptions, the coefficient of a feature $\beta_j$ determines its significance and can be as the testing statistic (Section \ref{section classical significance test}).
% However, since neural networks are highly nonlinear and complex, there doesn't exist a direct testing statistic reflecting the significance of a feature.
To test the significance of a feature $x_j$, we first need a reasonable measure as the testing statistic to represent the relationship between $x_j$ and $y$.
\textit{n}FBST is a flexible framework that can be applied to global significance, local significance, and instance-wise significance testing problems.
The weighted average of the partial derivative, with weights defined by a positive measure $\mu$, is adopted as the testing statistic in \cite{horel2020significance}:
\begin{equation}\label{equation testing statistic weighted average derivative}\small
    \eta(\theta)=\int_{\mathcal{X}}\left(\frac{\partial f_0(x)}{\partial x_j}\right)^2 d \mu(x).
\end{equation}
This reflects the global significance of the overall data, whose value will not change when the data distribution is fixed.
In non-linear contexts, the significance of a feature in sub-population distribution
% for a single instance or a subset of data 
is dynamic and varies with its range.
We consider a simple case $f_0(X)=\operatorname{ReLU}(x_0)$ where $x_0 \sim \mathcal{N}(0, 1)$.
It can be calculated that $\eta(\theta)=\frac{1}{2}$, which is a constant without considering the specific values of $x_0$.
% \begin{equation}\label{equation testing statistic weighted average derivative value}
%     \eta(\theta)=\int_{\mathcal{X}}\left(\frac{\partial f_0(X)}{\partial x_0}\right)^2 d \mu(X) = \frac{1}{2}.
% \end{equation}
If we define $\mathcal{X}_i \subseteq \mathcal{X}$, the local significance testing statistic value is dynamic as $\mathcal{X}_i$ varies, that is
% $\mathcal{X} = \mathcal{X}_1 \cup \dots \cup \mathcal{X}_k$ satisfying $\mathcal{X}_i \cap \mathcal{X}_j = \phi$ for $\forall 1 \le i, j \le k, i \neq j$, 
\begin{equation}\label{equation testing statistic weighted average derivative partial}\small
    \eta(\theta, \mathcal{X}_i)=\int_{\mathcal{X}_i}\left(\frac{\partial f_0(X)}{\partial x_j}\right)^2 d \mu(X).
\end{equation}
In this example, if we define $\mathcal{X}_1=\{X: x_0 < 0\}, \mathcal{X}_2=\{X: x_0 \geq 0\}$, we obtain $\eta(\theta, \mathcal{X}_1) = 0, \eta(\theta, \mathcal{X}_2) = \frac{1}{2}$.
% \begin{equation}\label{equation testing statistic weighted average derivative separate value}
%     \eta(\theta, \mathcal{X}_i)=\int_{\mathcal{X}_1}\left(\frac{\partial f_0(X)}{\partial x_0}\right)^2 d \mu(X) = 0,
%     \eta(\theta, \mathcal{X}_i)=\int_{\mathcal{X}_2}\left(\frac{\partial f_0(X)}{\partial x_0}\right)^2 d \mu(X) = \frac{1}{2}.
% \end{equation}
It is clear that under partial derivative settings, $x_0$ is insignificant when its value is less than zero, but significant when its value is greater than zero.
Further, $\mathcal{X}_i$ can contain only one data, that is the instance-wise significance testing statistic
\begin{equation}\label{equation testing statistic weighted average derivative instance}\small
    \eta(\theta, x_j)=\frac{\partial f_0(X)}{\partial x_j}.
\end{equation}

\textit{n}FBST is a general framework and supports significance testing based on various feature importance measures.
In our implementation, we select LRP\cite{binder2016layer}, LIME\cite{ribeiro2016should}, and DeepLIFT\cite{shrikumar2017learning} as testing statistics and the corresponding methods are called LRP-\textit{n}FBST, LIME-\textit{n}FBST, and DeepLIFT-\textit{n}FBST respectively.
Eq (\ref{equation testing statistic weighted average derivative instance}) uses gradient as the testing statistic and we name it Grad-\textit{n}FBST.
% These methods have been introduced in related works (Section \ref{section related works}) and the testing methods based on these testing statistics are called

% \begin{itemize}[leftmargin=5pt]
%     \item \textbf{Gradient \cite{simonyan2013deep} or Saliency.} 
%     It is a typical backpropagation-based method that uses the gradient of the output w.r.t. inputs as the feature importance.
%     \item \textbf{LRP \cite{binder2016layer} or Gradient $\times$ Input.} 
%     It is a gradient-based method. \cite{kindermans2016investigating} shows the LRP rules for ReLU networks are equivalent within a scaling factor to gradient $\times$ input in some conditions.
%     \item \textbf{DeepLIFT \cite{shrikumar2017learning}.} It is a backpropagation-based method by backpropagating the contributions of all neurons calculated by the difference between activation and reference.
%     \item \textbf{LIME \cite{ribeiro2016should}.} It is a perturbation-based method, using the data collected by perturbing near sample points to construct a local linear model.
% \end{itemize}

For global significance, Eq (\ref{equation testing statistic weighted average derivative}) is difficult to capture enough information when the distribution of the testing statistic is complex.
Therefore, we propose a Quantile-based Global Significance, namely Q-GS.
First, we sort all Bayesian evidence of instance-wise significance in descending order.
Then, we set a threshold $\lambda$ and select the quantile of the sorted evidence.
It satisfies that the percentage of evidence over it reaches the threshold $\lambda$.
% For example, if we set the threshold $0.05$, then we need to find out the greatest number and ensure not more than $0.05$ of the Bayesian evidence are less than it.

% \subsection{Convergence of Bayesian Factor}

% \textbf{TODO.}

\section{Experiments}

\subsection{Toy Example}\label{section toy example}

\begin{figure*}[t]
    % \captionsetup[subfigure]{labelformat=empty}
    \centering
    \subfloat[Scatter of Evidence]{\includegraphics[width=1.3in]{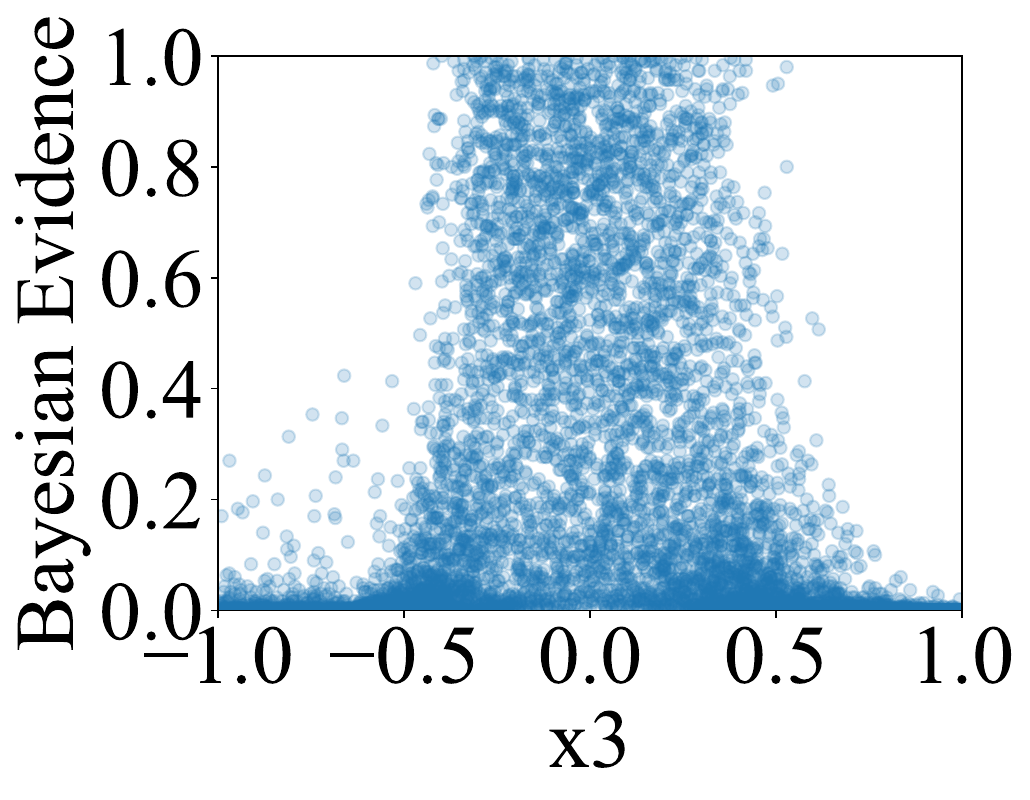}\label{gradient_scatter}}
    \subfloat[$0 \le |x_3| < 0.25$]{\includegraphics[width=1.3in]{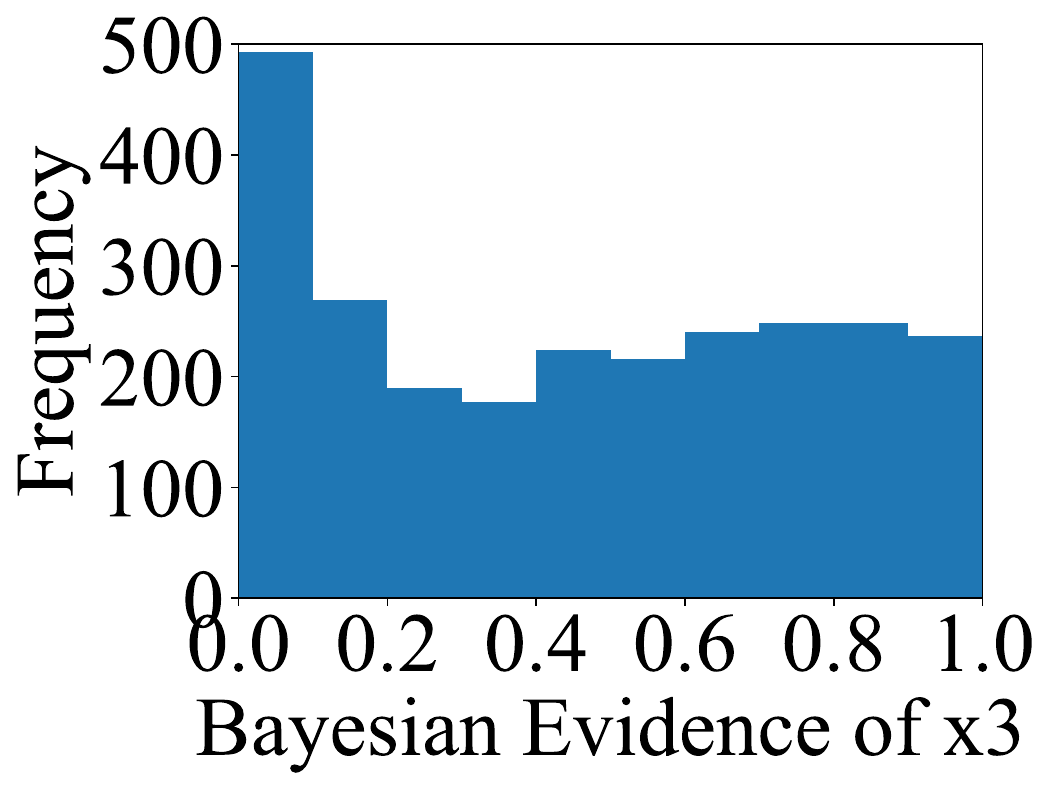}\label{gradient_0_0.25}}
    \subfloat[$0.25 \le |x_3| < 0.5$]{\includegraphics[width=1.3in]{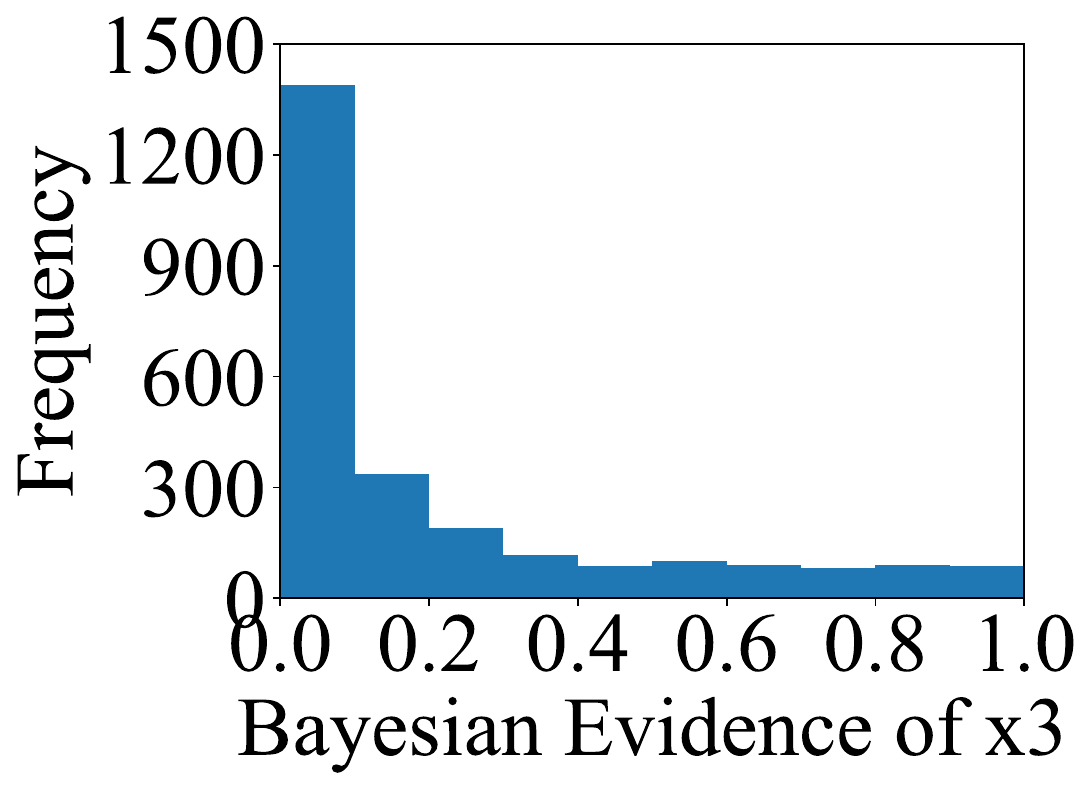}\label{gradient_0.25_0.5}}
    \subfloat[$0.5 \le |x_3| < 0.75$]{\includegraphics[width=1.3in]{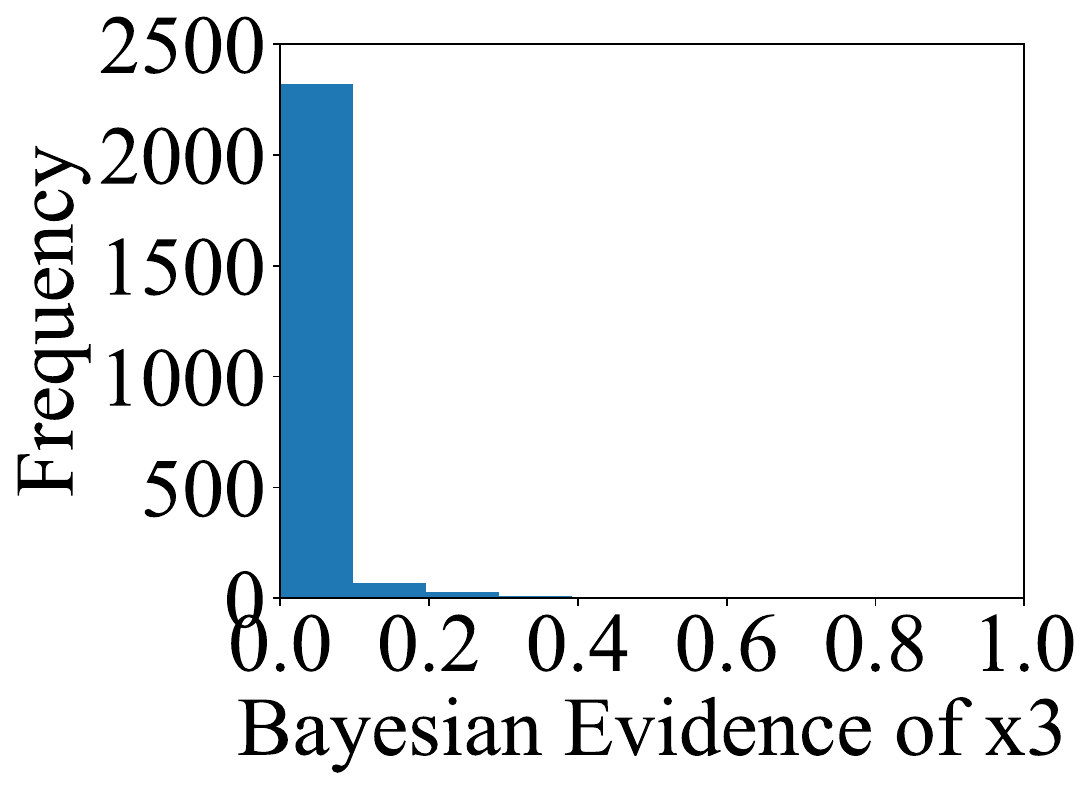}\label{gradient_0.5_0.75}}
    \subfloat[$0.75 \le |x_3| < 1$]{\includegraphics[width=1.3in]{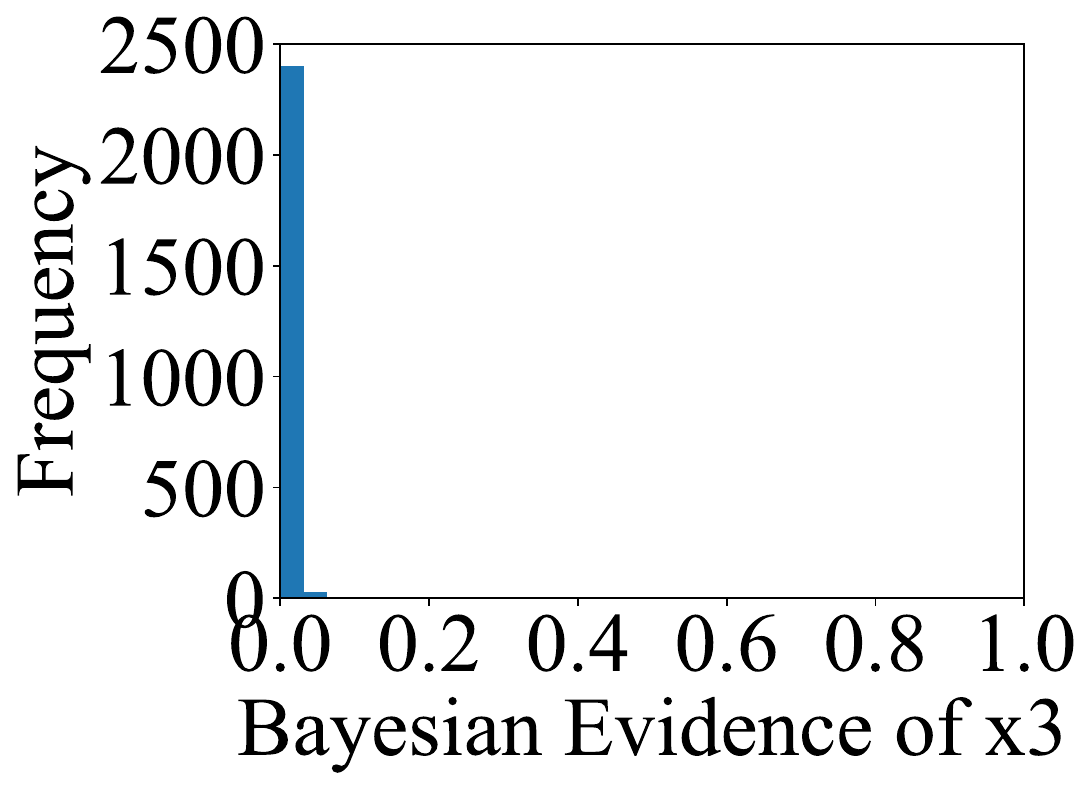}\label{gradient_0.75_1}}
    \caption{Bayesian evidence of $x_3$ obtained by Grad-\textit{n}FBST under different intervals on the toy example.}
    \label{figure toy example}
\end{figure*}

\begin{table*}[t]
  \centering
  \small
  \begin{tabular}{ccccccc}
    \toprule
     & \multicolumn{3}{c}{\textit{p-value}} & \multicolumn{3}{c}{Q-GS($\lambda=0.5$) by \textit{n}FBST} \\
    \cmidrule(r){2-4} \cmidrule(r){5-7}
    {Feature} & Bootstrap     & \makecell{likelihood\\ ratio test} & \textit{t}-test     & \makecell{Grad- \\ \textit{n}FBST} & \makecell{DeepLIFT-\\ \textit{n}FBST} & \makecell{LRP-\\ \textit{n}FBST}\\
    \midrule
    $x_0$ & $<$ 0.001      & 0.696             & 0.442             &  $<$0.001        & $<$0.001           & $<$0.001        \\
    % & $<$0.001 \\
    $x_1$ & $<$0.001      & 0.063             & 0.592             & 0.021          & $<$0.001           & 0.02        \\  
    % & 0.167 \\
    $x_2$ & $<$0.001      & 0.087             & 0.598             & 0.029          & $<$0.001           & 0.027         \\
    % & 0.197 \\
    $x_3$ & 0.027       & \textbf{0.813}    & 0.838             & 0.011          & 0.043            & 0.01          \\
    % & 0.93 \\
    $x_4$ & $<$0.001      & 0.209             & 0.604             & 0.005          & 0.007            & 0.003         \\
    % & 0.833 \\
    $x_5$ & $<$0.001      & 0.361             & 0.559             & 0.006          & 0.01             & 0.007         \\
    % & 0.77 \\
    $x_6$ & $<$0.001      & 0.318             & $<$0.001            &  0.278         & 0.203            & 0.273         \\
    % & 0.987 \\
    $x_7$ & \textbf{0.383} & 0.049          & \textbf{0.922}    & \textbf{0.637} & \textbf{0.637}   & \textbf{0.617}    \\
    % & \textbf{0.997} \\
    \bottomrule
  \end{tabular}
  \caption{Global significance testing results of different algorithms on the toy example, the maximum values are bolded.}
  \label{table different testing algorithms of variables on simulation_v3}
\end{table*}

We consider the following data generation process
\begin{equation}\label{equation toy example}\small
    y = 8 + x_0^2 + x_1x_2 + \cos(x_3) + \exp(x_4x_5) + 0.1x_6 + 0x_7 + \epsilon,
\end{equation}
where $X = [x_0, x_1, \dots, x_{7}] \sim \mathcal{U}(-1, 1)^{8}, \epsilon \sim \mathcal{N}(0, 1)$.
The variable $x_7$ has no influence on $y$.
Our goal is to differentiate $x_7$ from other features, that is to determine $x_7$ as insignificance but others as significance.
We compare the following three classical testing methods as the baselines:
\begin{itemize}[leftmargin=15pt]
    \item \textbf{Bootstrap \cite{Bradley1979bootstrap}.}
    It gets the distribution of the testing statistic from samples repeatedly drawn from the original data and simulates the mean and variance of the population to perform \textit{Z}-test.
    \item \textbf{Likelihood ratio test \cite{fisher1922interpretation}.}
    By training an unconstrained model incorporating all variables and a nested model with restricted variables, and comparing their likelihoods, we will obtain the standard asymptotic chi-square distribution if the unconstrained model is assumed correctly.
    \item \textbf{\textit{t}-test for linear models \cite{student1908probable}.}
    It calculates the estimated coefficient divided by its standard error and tests the result whether to follow the \textit{t}-distribution.
\end{itemize}
From table \ref{table different testing algorithms of variables on simulation_v3} we have the following observations:
\begin{itemize}[leftmargin=15pt]
    \item First, for the three classical testing methods, only Bootstrap accurately identifies the global significance of all features.
    Usually, we set the significance level $\alpha$ and compare \textit{p-value} with it.
    If the \textit{p-value} is smaller than the significance level, we reject the null hypothesis and accept the alternative hypothesis.
    That is, the smaller the \textit{p-value}, the more we can determine a feature is significant.
    If we set $\alpha=0.05$, the \textit{p-value} of Bootstrap satisfies that only $x_7$ is greater than $\alpha$ but others are not.
    However, the likelihood ratio test and \textit{t}-test can hardly distinguish correctly.
    This is probably because their assumptions about $f_0$ are too strong and lead to errors.
    \item Second, all \textit{n}FBST methods based on different testing statistics perform well.
    Here, we set $\lambda=0.5$ to obtain Q-GS.
    It means only when more than half of the instance-wise testing results reflect insignificance, we determine the feature as insignificant globally.
    The smaller the Q-GS, the less the evidence for $H_0$, and we tend to reject that the feature is insignificant.
    The results show all \textit{n}FBST methods provide strong evidence about $x_7$ for $H_0$ but little evidence for other features.
    \item Third, instance-wise significance can provide us with more insights than global significance.
    For the likelihood ratio test and \textit{t}-test methods, \textit{p-value} for $x_3$ is high even the highest.
    We plot scatters of the evidence for $x_3$ obtained by \textit{n}FBST and plot histograms under different $x_3$ intervals.
    As shown in Figure \ref{figure toy example}, its evidence of Grad-\textit{n}FBST is more concentrated on one when the value of $x_3$ is close to zero.
    This is consistent with Eq (\ref{equation toy example}) as ${\partial f_0(X)}/{\partial x_3}=-\sin(x_3)$.
    % We infer that the inconsistent performance of the three classical testing methods may be due to averaging different situations together and global significance is more coarse-grained than instance-wise significance.
    We conclude that global significance is more coarse-grained than instance-wise significance due to averaging different situations together.
\end{itemize}

% \begin{figure}
%   \centering
%   \includegraphics[width=5.5in]{figure_toy_example.png}
%   \caption{Bayesian evidence of $x_3$ obtained by Grad-\textit{n}FBST(a), DeepLIFT-\textit{n}FBST(b), LRP-\textit{n}FBST(c) under different values of $x_3$.}
%   \label{figure toy example}
% \end{figure}

\subsection{Simulation Experiments}

\begin{table*}[t]
  \centering
  \small
  \begin{tabular}{ccccccccc}
    \toprule
    Dataset & Metric & Bootstrap     & \makecell{likelihood\\ratio test} & \textit{t}-test     & \makecell{Grad- \\ \textit{n}FBST} & \makecell{DeepLIFT-\\ \textit{n}FBST} & \makecell{LRP-\\ \textit{n}FBST} & \makecell{LIME-\\ \textit{n}FBST} \\
    \midrule
    \multirow{3}[1]{*}{Dataset 1} 
    & Precision & 0.54      & 0.91             & 0.96             & 1          & 1           & 1         & 1 \\
    & Recall    & 0.98      & 0.80             & 0.86             & 1          & 1           & 1         & 1 \\
    & F1-score  & 0.70      & 0.85             & 0.91             & 1          & 1           & 1         & 1 \\
     \midrule
    \multirow{3}[1]{*}{Dataset 2} 
    & Precision & 0.55      & 0.94             & 0.91             & 1          & 1           & 1         & 1 \\
    & Recall    & 0.98      & 0.66             & 0.86             & 1          & 1           & 1         & 1 \\
    & F1-score  & 0.71      & 0.78             & 0.89             & 1          & 1           & 1         & 1 \\
    \midrule
    \multirow{3}[1]{*}{Dataset 3} 
    & Precision & 0.51      & 0.87             & 0.90             & 0.82          & 0.80           & 0.83         & 0.85 \\
    & Recall    & 1      & 0.54             & 0.54             & 0.84          & 0.90           & 0.86         & 0.80 \\
    & F1-score  & 0.68      & 0.67             & 0.68             & 0.83          & 0.85           & 0.84         & 0.82 \\
    \bottomrule
  \end{tabular}
  \caption{Precision, Recall and F1-score for global significance on different Datasets.}
  \label{table precision and recall of global significance on dataset 1 and dataset 2}
\end{table*}

\begin{figure*}[t]
  \centering
  \includegraphics[width=6in]{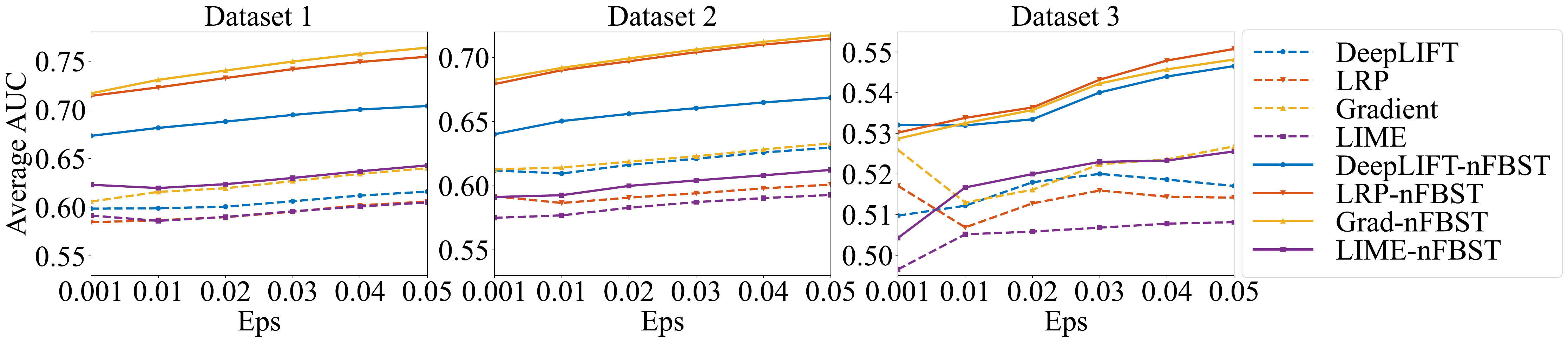}
  \caption{Average AUC of all features for instance-wise significance before and after \textit{n}FBST under different eps.}
  \label{figure auc}
\end{figure*}

In this section, we conduct experiments on three simulation datasets for analysis.
Compared to three classical testing methods, we found \textit{n}FBST can perfectly distinguish the global significance while others cannot.
Compared to five feature importance methods, we found \textit{n}FBST can improve the ability to test instance-wise significance.

\subsubsection{Data Generation Process}

We consider the input features $X = [x_0, x_1, \dots, x_{99}] \sim \mathcal{U}(-1, 1)^{100}$ and the data generation process
\begin{equation}\label{data-generating 1}\small
  \begin{aligned}
    y = f_0(X) + \epsilon,
  \end{aligned}
\end{equation}
where $\epsilon \sim \mathcal{N}(0, 0.01)$ and $f_0$ is a neural network function whose weights and biases are initialized randomly.
Then only the last fifty features are insignificant by setting the corresponding weights to zero.
Our goal is to select these fifty insignificant features accurately.

According to the above generation process, we generate two sets of 10,000 independent samples, namely Dataset 1 and Dataset 2.
The difference between them is that the structure of $f_0$ for Dataset 1 is three hidden layers of 20 nodes but three hidden layers of 16 nodes for Dataset 2.
In our experiments, we only adopt the structure with three hidden layers of 20 nodes as the trained model to simulate conditions where it has the same or different structure from $f_0$.
Then, we reduce the data size to one-tenth of the Dataset 2, that is 1,000 independent samples, namely Dataset 3, to simulate the scenario of small data.

\subsubsection{Test the Global Significance}

% First, we compare the performance of the global significance of \textit{n}FBST with three classical testing methods, Bootstrap, likelihood ratio test, and \textit{t}-test for linear models.
For the global significance of each feature, there are two possible testing results, insignificant or significant.
Therefore, we can consider the task of testing as a binary classification problem.
Specifically, significant represents the positive class, and insignificant represents the negative class.
% TP, TN, TN, and FN can be calculated further.
TP means identifying significant features correctly.
TN means identifying insignificant features correctly.
FN means misidentifying a feature that should be significant as an insignificant feature.
FP means misidentifying a feature that should be insignificant as a significant feature.
Precision=TP/(TP+FP), reflects the accuracy of correctly testing significant features.
Recall=TP/(TP+FN), reflects the completeness of correctly identifying significant features.
F1-score combines precision and recall to calculate the harmonic mean.

Table \ref{table precision and recall of global significance on dataset 1 and dataset 2} shows these metrics on three simulation datasets compared to three classical testing methods.
As with the settings of ``Toy Example'', we set the significance level for classical testing methods as 0.05 and $\lambda=0.5$ to obtain Q-GS.
First, Bootstrap tends to determine a feature as significant thus resulting in high recall but poor precision.
On the contrary, the likelihood ratio test tends to determine a feature as insignificant thus resulting in poor recall but fine precision.
If we compare comprehensively, \textit{t}-test method outperforms the two methods with the highest F1-score.
Second, all \textit{n}FBST methods based on different testing statistics perform perfectly, with the F1-score of 1.
Compared to the classical testing methods, the improvement is largely due to the flexible hypothesis of \textit{n}FBST, as neural networks can fit more complex cases of $f_0$.

\subsubsection{Test the Instance-wise Significance}

% Testing the significance of features locally means determining whether a feature is significant or not for a particular prediction.
% We summarize \textit{n}FBST into two main advantages based on the case study and simulation experiments.
% \begin{itemize}
%     \item \textit{n}FBST overcomes the limitation that most existing point-estimation methods only consider the maximum probability of occurrence.
%     By evaluating all occurrences of the overall distribution, \textit{n}FBST can obtain more comprehensive results.
%     \item Feature importance analysis assigns a score for a prediction, which shows the significance of the feature.
%     After combining \textit{n}FBST, the ability to distinguish significant and insignificant features has been improved further, for all the four methods tested.
% \end{itemize}

% \begin{figure}
%   \centering
%   \includegraphics[width=4.5in]{figure_auc_dataset1.png}
%   \caption{Comparison of AUC for each feature before and after \textit{n}FBST on Dataset 1.}
%   \label{figure auc dataset1}
% \end{figure}

For the instance-wise significance of each feature, there are also two possible testing results, insignificant or significant.
Specifically, there are 10,000 instances of 100 features for Dataset 1 and Dataset 2.
As the last fifty features are insignificant by setting the corresponding weights to zero, our focus is mainly on the first fifty features.
Because the instance-wise significance of these features varies with their values.
First, we calculate the gradients of $f_0$ on each instance and adjust different precision thresholds, namely $\text{eps}$, to label the instance-wise significance.
If the gradient is less than eps, we label it insignificant, otherwise significant.
Then, we evaluate the performance by ROC and AUC.
Most existing testing methods don't distinguish global significance and instance-wise significance and only focus on the former.
Therefore, we select feature importance analysis methods as baselines.
They assign a feature importance score for a prediction, which reflects the significance of the feature learned from the model.

From Figure \ref{figure auc}, we have the following observations:
\begin{itemize}[leftmargin=15pt]
    \item First, diverse measure-based \textit{n}FBST consistently surpasses primary feature importance methods across various epsilon settings. 
    Through the comparison of Grad and Grad-\textit{n}FBST, DeepLIFT and DeepLIFT-\textit{n}FBST, LRP and LRP-\textit{n}FBST, LIME and LIME-\textit{n}FBST, we infer that the integration of \textit{n}FBST enhances the capacity to discern instance-wise significant and insignificant features.
    % \item First, different measure-based \textit{n}FBST outperforms primary feature importance methods under different eps settings.
    % By comparing Grad and Grad-\textit{n}FBST, DeepLIFT and DeepLIFT-\textit{n}FBST, LRP and LRP-\textit{n}FBST, LIME and LIME-\textit{n}FBST respectively, we infer that the ability to distinguish instance-wise significant and insignificant features has been improved after combining \textit{n}FBST.
    % Moreover, it can be concluded that our method isn't influenced by the precision of ground truth.
    % The specific AUC of each feature under different eps is shown in the Appendix.
    Moreover, it can be concluded that our approach remains unaffected by the precision of the ground truth.
    The distinct AUC for each feature under varying epsilon values is presented in the Appendix.
    \item Second, LIME and LIME-\textit{n}FBST perform worse than other methods.
    That's because LIME is a perturbation-based method that constructs a local linear model based on the data collected by perturbing near sample points.
    Its performance is limited by sampling efficiency.
    % It is different from other back-propagation-based methods, which backpropagate gradients (or variation of gradients) or reference scores to obtain feature importance.
    % The biggest problem of LIME is the computational efficiency because each sampling requires a forward propagation and this process is hard to accelerate.
    % Besides, its performance is related to the number of samples, because more samples lead to a more accurate linear model somehow.
    % In our experiment, we sample 100 times for each calculation and we suspect that this constrains its performance.
\end{itemize}

% Testing the significance of features globally means determining whether a feature is significant or not in the whole data.
% This problem is consistent with local significance under linear assumptions, and most testing methods only test globally actually.

\subsection{Real World Experiments}

\begin{figure*}[t]
    % \captionsetup[subfigure]{labelformat=empty}
    \centering
    \subfloat[$x_8=0$]{\includegraphics[width=1.2in]{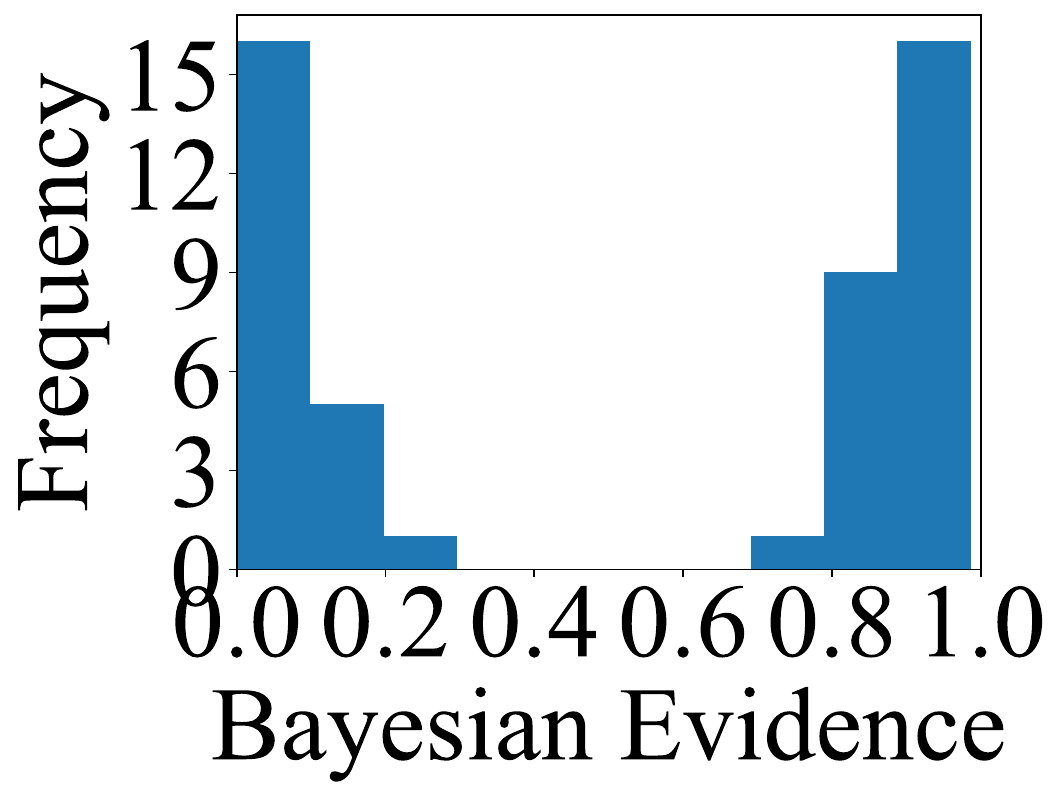}}
    \subfloat[$x_8=1$]{\includegraphics[width=1.2in]{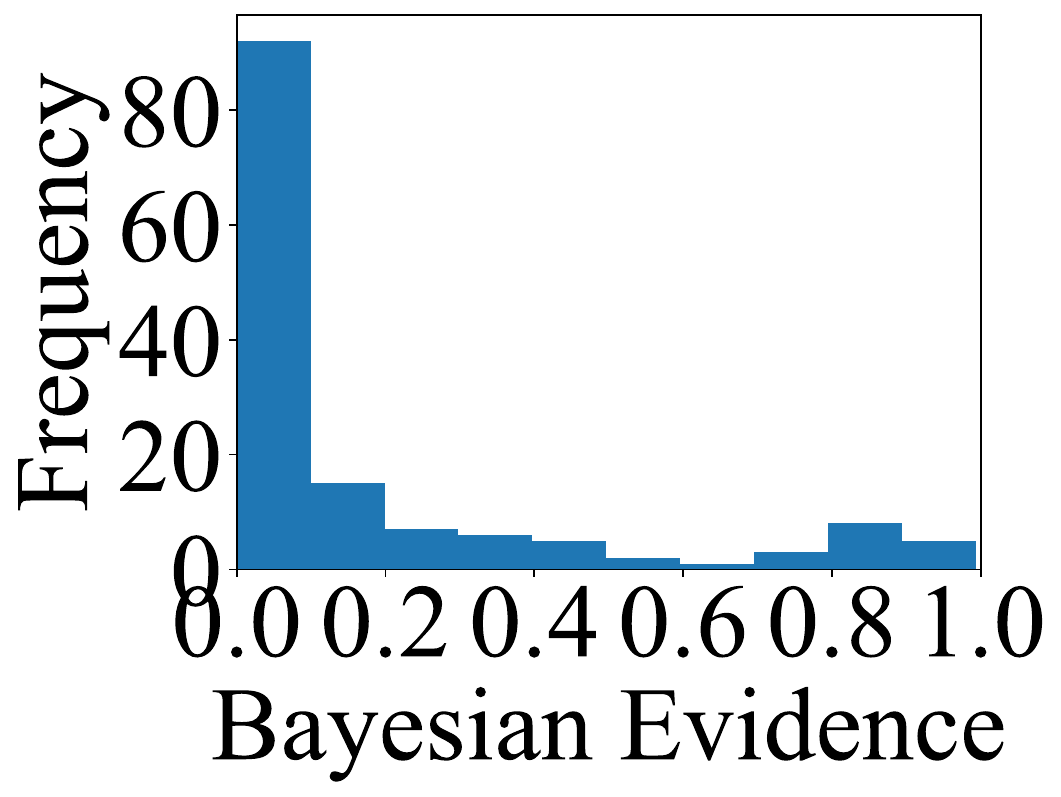}}
    \subfloat[$x_8=2$]{\includegraphics[width=1.2in]{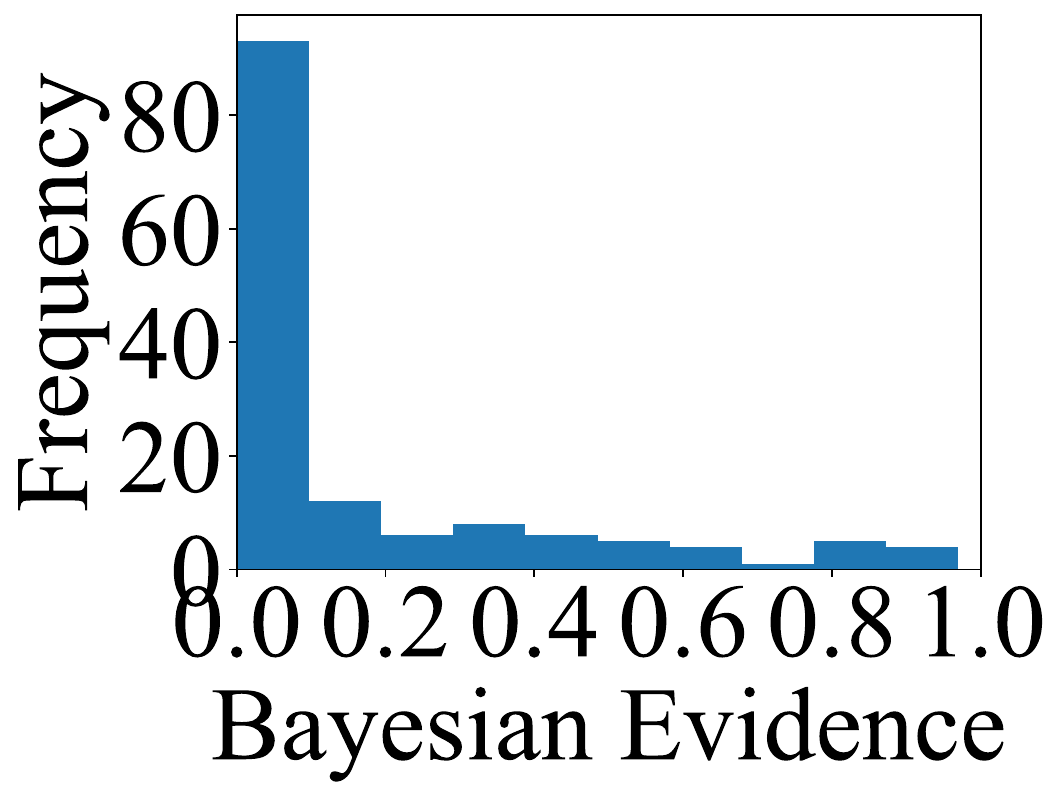}}
    \subfloat[$x_8=3$]{\includegraphics[width=1.2in]{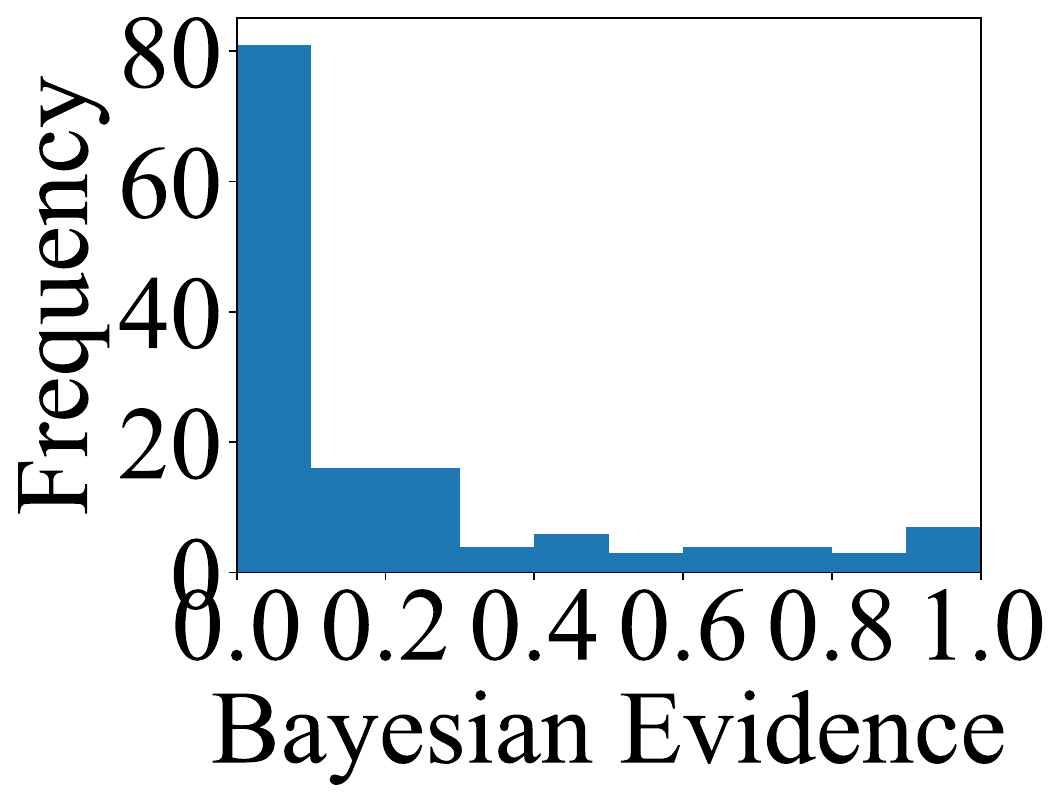}}
    \subfloat[$x_8=4$]{\includegraphics[width=1.2in]{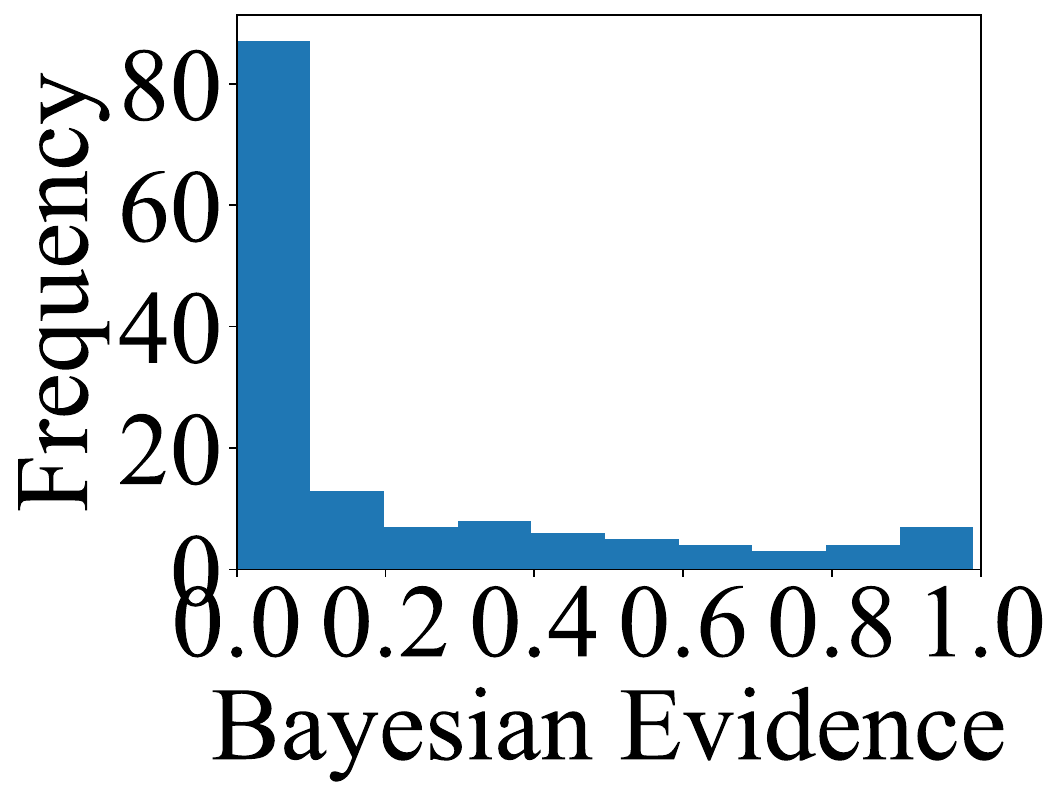}}
    \subfloat[$x_8=5$]{\includegraphics[width=1.2in]{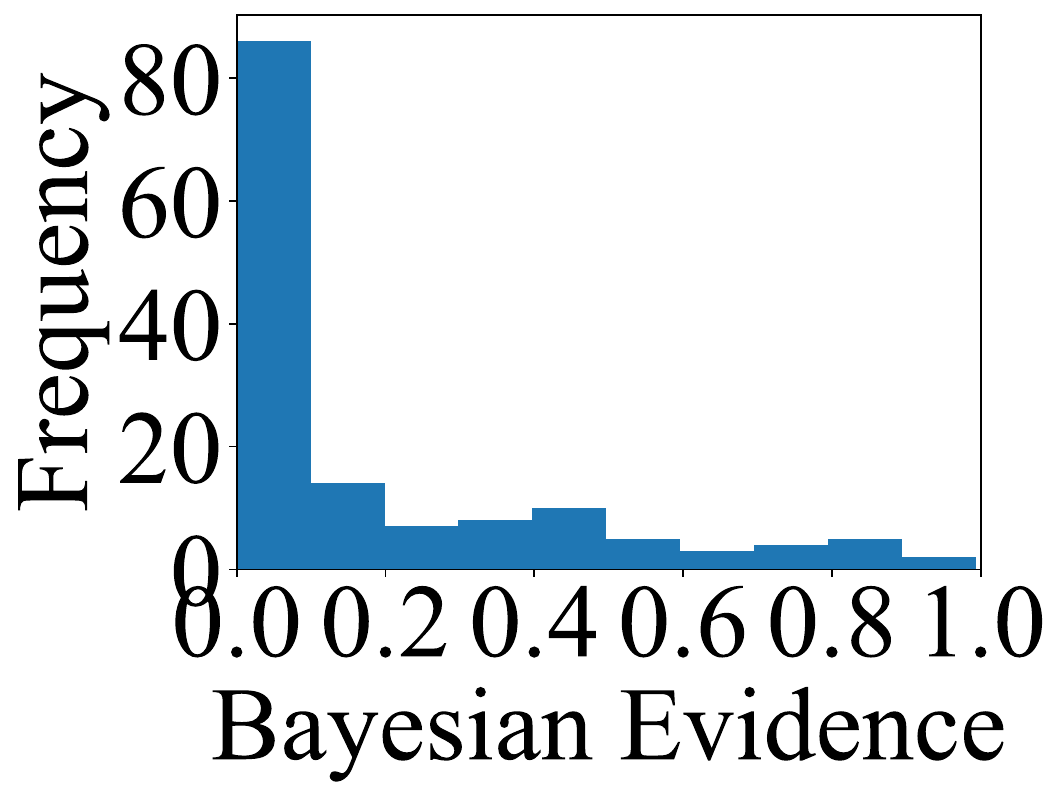}}
    \caption{Histograms of gradient distributions for different values of $x_8$ on energy efficiency dataset.}
    \label{figure X8 testing results}
\end{figure*}

In this section, we analyzed the performance of \textit{n}FBST in real world scenarios through UCI and image datasets.
On energy efficiency, we focused on analyzing feature $x_8$ and found the instance-wise testing results are consistent with its physical truth.
On MNIST, by comparing different feature importance methods, we found \textit{n}FBST recognized the object information in the image more prominently.

% \subsubsection{Residential Energy Efficiency Evaluation}

The energy efficiency dataset comprises 768 samples and 8 features.
It aims to predict the dependent target $y$ (HL, heating load), which determines the specifications of the heating equipment needed to maintain comfortable indoor air conditions.
The descriptions of the features and target are shown in the Appendix.
% Table \ref{table different testing algorithms of variables on energy efficiency} shows that t-test, likelihood-ratio test, and different measure-based \textit{n}FBST all identify $X_6$ and $X_8$ as insignificant, which is consistent with \cite{tsanas2012accurate}.
% But Bootstrap fails to distinguish $X_6$ and $X_8$ with other features.
% Note that the smaller the p-value or the greater the Bayesian evidence by \textit{n}FBST, the lower probability of $H_0$ occurring.
% DeepLIFT-FBST and LRP-FBST also recognized $X_3$ as insignificant, that's because we standardized the data to fit the model well, causing some original values of $X_3$ to be 0.
From figure \ref{figure X8 testing results}, we find that testing results are more concentrated around one when $x_8$ equals zero, while others are not.
It indicates that the instance-wise significance of $x_8$ is different under different values, insignificant if its value is zero.
The research in \cite{tsanas2012accurate} confirms our findings.
There are six possible values for $x_8$ (0, 1, 2, 3, 4, 5) in total.
% The problem of testing the significance of features becomes much more complicated under non-linear assumptions.
% If we analyze more deeply why $X_8$ is judged to be insignificant, we will find out there are six total possible values for it.
When $x_8$ equals zero, it means no glazing areas and that's why $x_8$ doesn't make sense in this situation.
% It is delightful if we could make precise judgments according to different values of the feature.
% In other words, testing whether a feature is significant or not overall is coarse-grained but testing locally at each specific data is the most find-grained.
% That is exactly what \textit{n}FBST is focused on.
In conclusion, \textit{n}FBST can effectively discover instance-wise significance in real world data.

\begin{figure}[t]
  \centering
  \includegraphics[width=3in]{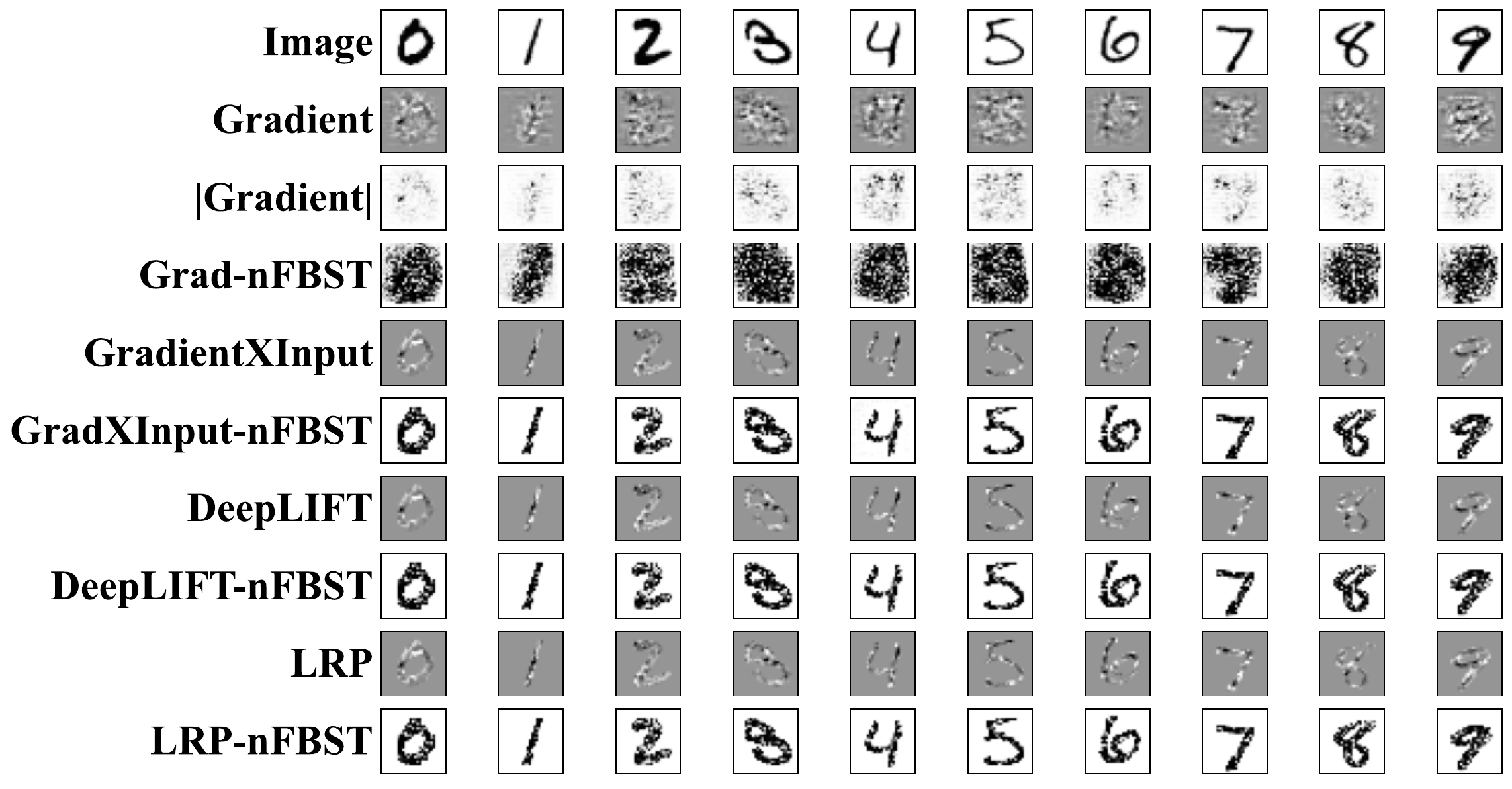}
  \caption{Visualization of scores calculated by different methods for the target class.}
  \label{figure mnist}
\end{figure}

The testing problem for MNIST is defined as testing each pixel of a digit image and distinguishing significant pixels from insignificant pixels for the target.
It involves a weakly supervised semantic segmentation task in computer vision.
Ideally, the pixels related to the target class should be assigned higher scores than the background pixels.
For feature importance analysis methods, they generate a saliency map based on feature importance scores.
For \textit{n}FBST, Bayesian evidence represents the evidence supporting $H_0$, which is pixel-wise insignificance in this task.
Figure \ref{figure mnist} shows that the object pixels are more prominent after using \textit{n}FBST.
Besides, Grad-\textit{n}FBST identifies a messy area because the primary gradient method recognizes poorly.
When we multiply it with the input, the performance is improved and this problem doesn't exist in other methods (LRP and DeepLIFT).

\section{Related Works}\label{section related works}

% \subsection{interpretability of Deep Learning}

In recent years, there has been an increasing amount of literature on the interpretability of deep learning \cite{wang2019empowering, 10.1145/3292500.3330647, wang2020interpretability, 8834829, wang2022traffic, wang2021personalized, wang2019svm, cong2021alphaportfolio, ji2020interpretable, ji2022stden}, one of which is feature importance analysis.
% Explaining a black box model is also an important topic that could tell us the defects and advise us on how to improve the model.
% Our method \textit{n}FBST is extended based on the measures obtained from feature importance analysis (also known as sensitivity analysis) as well.
The first group propagates an importance score from the output neuron backward to the input.
Most of them are gradient-based, including saliency map\cite{simonyan2013deep}, deconvolution\cite{zeiler2014visualizing}, guided backpropagation\cite{springenberg2014striving} and integrated gradients\cite{sundararajan2017axiomatic}.
The first three methods have different strategies to calculate the gradient when passing through the ReLU layer, but cannot show negative contributions and face discontinuity.
The integrated gradients method increases the computational cost by computing the integral.
% \cite{simonyan2013deep} first proposes using the gradient of the output w.r.t. pixels of an input image to compute a “saliency map” of the image.
% Later, \cite{zeiler2014visualizing} extends the idea to the convolutional neural networks by using a deconvolutional network.
% to analyze the high-level units and backpropagate their activation to the input to identify the important parts of the input image responsible for activation.
% Then, guided backpropagation combined these two approaches \cite{springenberg2014striving}.
% Essentially, all three methods calculate the derivative of the input through back-propagating, but different from the strategies for handling the gradient when passing through the ReLU layer.
% However, since functions like ReLU have a negative gradient of zero, they cannot show negative contributions and have the problem of discontinuity.
% Therefore, the integrated gradients method is proposed, which computes the integral of the gradient of the input scaled from some starting value to the current value \cite{sundararajan2017axiomatic}.
% \cite{sundararajan2017axiomatic} proposed the integrated gradients method which computes the integral of the gradient of the input scaled from some starting value to the current value, but increases the computational cost.
Another approach is the Layer-wise Relevance Propagation proposed by \cite{binder2016layer}.
\cite{kindermans2016investigating} shows that the LRP rules for ReLU networks are equivalent within a scaling factor to gradient $\times$ input in some conditions.
Moreover, DeepLIFT \cite{shrikumar2017learning} and SHAP \cite{lundberg2017unified} do not compute gradients but are also based on back-propagation.
% SHAP is a unified framework with a set of desirable properties for interpreting predictions, including DeepLIFT.
% In our experiments, we select it as our baseline.
In contrast, the second group of methods makes perturbations to individual inputs or neurons\cite{zeiler2014visualizing}.
%and observe the impact on later neurons in the network.
% The first work is \cite{zeiler2014visualizing}, which occludes different segments of an input image and visualizes the change.
% in the activation of later layers.
A typical approach is LIME \cite{ribeiro2016should}, where data are collected by perturbing near sample points to construct a local linear model.
However, it is computationally expensive and requires a large number of samples to obtain reliable results.
% The biggest problem of LIME is the computational efficiency because each sampling requires a forward propagation and this process is hard to accelerate.
% Besides, its performance is related to the number of samples, because more samples lead to a more accurate linear model somehow.
% Besides, Kernel SHAP proposed by \cite{lundberg2017unified} also falls into this category and is used as a baseline as well.
%  However, such methods can be computationally inefficient because of a separate forward propagation through the network for each perturbation.

% According to different criteria, these methods can be classified into different categories.
The above methods aim to explain the predictions of a model locally at a specific instance, while others aim to understand how the model works globally.
% In practical applications, providing a global interpretation of a deep learning model is usually more difficult than providing a local interpretation.
The partial dependence plot (PDP) shows the marginal impact of one or two features on the model prediction \cite{friedman2001greedy}\cite{greenwell2018simple}.
% \cite{greenwell2018simple} proposes a simple partial dependence-based feature importance measure and the basic motivation is that the greater the variation in PDP, the more important the feature is.
\cite{datta2016algorithmic} measures the impact by calculating the difference in the quantity of interest when the data is generated according to the true distribution and the hypothetical distribution designed deliberately.
SP-LIME extends LIME to global by selecting typical points\cite{ribeiro2016should}.
% to achieve interpretability from local to global.
% In our paper, we propose Q-GS to extend \textit{n}FBST from local to global.

% Unlike global interpretability, local interpretability of a model is oriented to the input sample and can usually be achieved by analyzing the contribution of each feature to the final decision outcome, also known as feature importance analysis or sensitivity analysis.

% Despite the variety of methods mentioned above, they all focus on assigning feature importance.
% There is also prior work treating the significance of variables in neural network regression models.
There is also prior work treating the significance of variables.
One is to regard neural networks as parametric formulations \cite{olden2002illuminating,white1989learning,white1989some,vuong1989likelihood} but restricts the model structure.
The testing statistic is not necessarily identifiable due to the non-identifiability of neural networks.
The other is to regard neural networks as nonparametric models \cite{gozalo1993consistent, lavergne1996nonparametric, yatchew1992nonparametric, fan1996consistent, lavergne2000nonparametric, racine1997consistent}.
However, most of them study in the context of kernel regressions and can be computationally challenging because of Bootstrap.
% The latest study is \cite{horel2020significance}, which proposes the significance test for neural networks with a single hidden layer by theoretical proof and experiments.
% However, due to the complexity of neural networks, it fails to extend to deep neural networks further, which is the target we try to address.
The latest related research restricts the model structure to a single hidden layer and only tests the global significance \cite{horel2020significance}.
% Despite the universal approximation theorem for neural networks \cite{hornik1989multilayer}, it is almost impossible to widen the hidden layer infinitely in practice.

\section{Conclusion}

In this paper, we propose to conduct the Full Bayesian Significance Testing for neural networks, called \textit{n}FBST.
It is a general framework that can be extended based on different measures.
To the best of our knowledge, we are the first to introduce significance testing into deep neural networks.
What's more, it offers a new perspective of exploring knowledge hidden behind the underlying relationship between features and targets in a rigorous way, rather than explaining the estimated relationship which contains estimation errors due to the randomness of the data generation process.
Extensive experiments on simulation and real-world datasets confirm the advantages of our proposed approach.

\section{Acknowledgments}

Prof. Wang’s work was supported by the National Natural Science Foundation of China (No. 72171013, 72222022, 72242101), the Fundamental Research Funds for the Central Universities (YWF-23-L-829) and the DiDi Gaia Collaborative Research Funds. Dr. He’s work was supported by the China National Postdoctoral Program for Innovative Talents (BX20230195).

\bibliography{aaai24}

\end{document}